\documentclass[runningheads]{llncs}

\usepackage[T1]{fontenc}
\usepackage{graphicx}
\usepackage{amsmath}
\usepackage{amssymb}
\usepackage{xcolor}
\usepackage{algorithm}
\usepackage{algpseudocode}
\usepackage{booktabs}
\usepackage{microtype}
\usepackage{hyperref}
\usepackage{cleveref}
\usepackage{subcaption}
\usepackage{makecell}
\usepackage{adjustbox}
\usepackage{stfloats}

\begin{document}

    \title{Adaptive Protection for Evolutionary Feature Construction in Symbolic Regression with Application to Credit Classification}
    \titlerunning{Evolutionary Feature Construction with Adaptive Protection}

    \authorrunning{H. Zhang et al.}
    \author{Hengzhe Zhang\inst{1} \and Qi Chen\inst{1} \and Bing Xue\inst{1} \and Lean Yu\inst{3} \and Wolfgang Banzhaf\inst{2} \and Mengjie Zhang\inst{1}}
    \institute{Centre for Data Science and Artificial Intelligence \& School of Engineering and Computer Science, Victoria University of Wellington, PO Box 600, Wellington 6140, New Zealand\\
    \email{\{hengzhe.zhang,qi.chen,bing.xue,mengjie.zhang\}@ecs.vuw.ac.nz}
    \and
    Department of Computer Science and Engineering, Michigan State University, East Lansing, MI 48824, USA\\
    \email{banzhafw@msu.edu}
    \and
    Business School, Sichuan University, Chengdu, 610065, China\\
    \email{yulean@amss.ac.cn}}

    \maketitle
    \setcounter{footnote}{0}

    \begin{abstract}
        Evolutionary feature construction has shown strong promise in symbolic regression by automatically discovering informative transformations of input features that enhance a simple base learner. However, existing approaches often lack explicit mechanisms to preserve important constructed features discovered during evolution, and valuable genetic material can be lost when genetic operators disrupt effective features. This paper introduces an adaptive protection mechanism that leverages feature importance metrics to selectively preserve constructed features during evolution. The mechanism provides stronger protection for more important constructed features while still allowing less important features to be modified and to incorporate useful building blocks from more important features. We evaluate the approach using multiple feature importance calculation methods and demonstrate its robustness across different base learners. Experimental results on 98 regression benchmark datasets show that the proposed mechanism consistently improves solution quality over baseline approaches, and experiments on two credit classification datasets demonstrate that the method also extends effectively to improve search effectiveness beyond symbolic regression.
    \end{abstract}

    \keywords{symbolic regression \and credit classification \and evolutionary computation \and feature construction \and feature importance \and genetic programming}

    \section{Introduction}

    Symbolic regression aims to discover mathematical expressions that best describe relationships in data, without assuming a predefined functional form~\cite{cava2021contemporary}. Formally, given a dataset $\mathcal{D} = \{(\mathbf{x}_i, y_i)\}_{i=1}^{n}$ where $\mathbf{x}_i \in \mathbb{R}^d$ are input features and $y_i \in \mathbb{R}$ are target values, symbolic regression seeks to find a function $f: \mathbb{R}^d \rightarrow \mathbb{R}$ that minimizes a loss function:
    $
    f^* = \arg\min_{f \in \mathcal{F}} \sum_{i=1}^{n} L(y_i, f(\mathbf{x}_i))
    $
    where $\mathcal{F}$ is the space of all possible mathematical expressions and $L$ is a loss function such as mean squared error.

\begin{figure}[!t]
        \centering
        \includegraphics[width=0.9\textwidth,trim=8pt 8pt 8pt 8pt,clip]{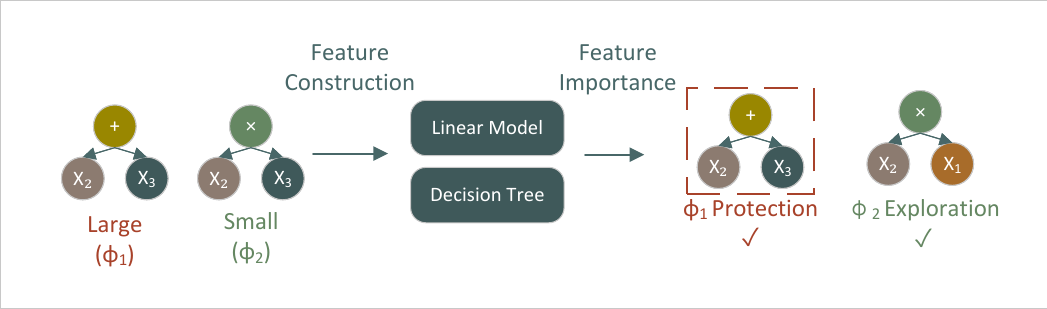}
        \caption{Universal protection framework that wraps around arbitrary genetic operators, analogously to depth-limit protection that is not bound to any specific genetic operator.}
        \label{fig:gpfc}
        \vspace{-4mm}
    \end{figure}

    Among symbolic regression methods, feature-construction-based symbolic regression is a popular approach that generates informative feature transformations $\phi_j: \mathbb{R}^d \rightarrow \mathbb{R}$, which are then used to build predictive models $f(\phi_1(\mathbf{x}), \allowbreak \ldots, \phi_k(\mathbf{x}))$~\cite{rovito2025interpretable}. Evolutionary feature construction-based symbolic regression represents a particularly promising variant where genetic programming (GP) automatically evolves these transformations~\cite{la2020learning,zhang2023modular}, creating a rich space of potential transformations that can be combined by a linear model or a decision tree, as illustrated in \Cref{fig:gpfc}.

    As with many other evolutionary approaches, a fundamental challenge in evolutionary feature construction lies in balancing the preservation of useful building blocks with the generation of new feature combinations. Genetic operators such as mutation and crossover can generate promising new constructed features, but they may also inadvertently disrupt valuable existing ones~\cite{banzhaf1998genetic}. This is problematic because modifying or removing an important feature can replace a highly useful building block with a less effective alternative, potentially degrading solution quality even when the offspring introduces other useful features.

    To address this challenge, feature importance metrics provide a natural mechanism to identify which constructed features contribute most to model performance~\cite{lundberg2017unified,kazemitabar2017variable}. Traditional approaches use model coefficients~\cite{zhang2023sr} and marginal contribution~\cite{behzadifar2025decompose}, while recent advances have introduced more sophisticated methods such as Shapley values~\cite{lundberg2017unified}. These metrics offer different perspectives on feature contribution and can guide the preservation of valuable genetic material during evolution. Although feature-importance-aware genetic operators have been developed~\cite{zhang2021correlation,zhang2025general}, there is no simple mechanism that is universally applicable to arbitrary genetic operators.

    This paper introduces an adaptive protection mechanism that uses feature importance to guide the preservation of good building blocks\footnote{\url{https://github.com/hengzhe-zhang/EvolutionaryForest/blob/master/evolutionary_forest/component/crossover/adaptive_feature_importance.py}}. As illustrated in \Cref{fig:gpfc}, this mechanism can be embedded in a universal protection framework that is compatible with any genetic operator. The key insight is that constructed features should receive protection proportional to their importance, so that highly useful features are preserved more reliably, whereas less important features remain free to be modified or replaced by genetic operators. In this way, feature protection preserves strong building blocks, while crossover and mutation continue to generate new feature combinations and replace less valuable ones.

    The specific objectives of this work are threefold:
    \begin{itemize}
        \setlength{\itemsep}{0pt}\setlength{\parskip}{0pt}
        \item We introduce an adaptive protection mechanism that protects constructed features using an importance-weighted score to guide restoration, providing stronger protection for more important constructed features while still allowing less important features to be modified or replaced.
        \item We compare multiple feature importance calculation methods across different model types to demonstrate the generality of our approach.
        \item We provide an empirical evaluation on symbolic regression benchmarks, together with an application to credit classification, demonstrating the effectiveness of the approach across different importance metrics and base learners.
    \end{itemize}

    \section{Related Work}\label{sec:related-work}

    \subsection{Evolutionary Feature Construction}

    Feature construction in evolutionary computation falls into three paradigms: wrapper, filter, and embedded. Wrapper methods~\cite{la2017ensemble} train a model on constructed features and use its performance as fitness. They include Multi-\allowbreak dimensional Multiclass Genetic Programming (M3GP)~\cite{munoz2019evolving}, Interaction-\allowbreak Transformation Evolutionary Algorithm (ITEA)~\cite{de2021interaction}, Multiple Regression Genetic Programming (MRGP)~\cite{arnaldo2014multiple}, and Genetic Programming Gene-\allowbreak pool Optimal Mixing Evolutionary Algorithm (GP-\allowbreak GOMEA)~\cite{virgolin2020explaining}. Filter methods~\cite{neshatian2012filter,tran2019genetic} use statistical or information-theoretic criteria, independent of the learner. Embedded methods~\cite{wang2024improving} integrate construction into the learning process. This work focuses on wrapper-based approaches, which have shown strong performance in symbolic regression and classification.

    \subsection{Building Block Analysis}

    The notion of building blocks---reusable, high-quality substructures---has long been studied in GP. Classical work focuses on structural building blocks and their role in bloat control~\cite{kinzett2009numerical}, whereas semantic building blocks characterize useful program fragments through their input--output behaviors~\cite{mcphee2008semantic}. Related work has also explored gene-pool optimal mixing with learned building blocks~\cite{virgolin2017scalable}, as well as building-block reuse in analog circuit synthesis~\cite{mcconaghy2011trustworthy}, large-scale Boolean problems~\cite{iqbal2013reusing}, and image classification~\cite{bi2023genetic}. More directly relevant, importance analysis has guided genetic operators at the tree level~\cite{zhang2024geometric}, subtree level~\cite{zhang2021correlation,nguyen2021automated}, and variable level~\cite{wang2025improving}. These methods rely on measures such as frequency analysis~\cite{virgolin2017scalable}, correlation analysis~\cite{zhang2021correlation}, and Shapley values~\cite{wang2025improving}, but crossover and mutation typically must be redesigned to exploit the identified components. It is therefore desirable to protect useful parts with an operator-agnostic mechanism that remains compatible with traditional operators, semantic operators~\cite{farinati2025study}, and LLM-based crossover~\cite{anthes2025transformer}.

    \section{Algorithm}\label{sec:algorithm}

    \subsection{Overall Workflow}\label{sec:workflow}

    \begin{figure}[!t]
        \centering
        \begin{subfigure}[b]{0.68\textwidth}
            \centering
            \includegraphics[width=\textwidth,trim=8pt 8pt 8pt 8pt,clip]{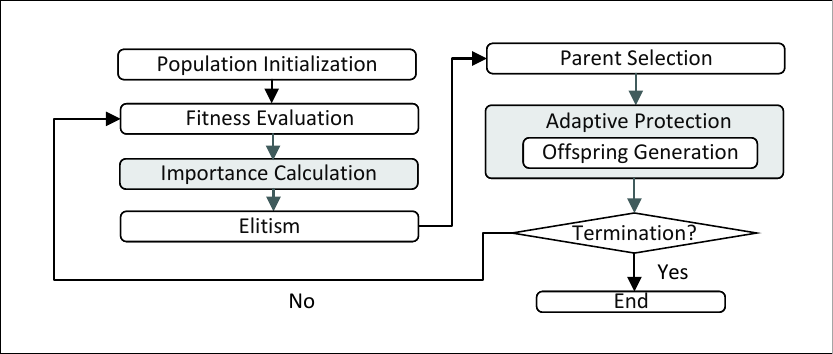}
            \caption{Workflow.}
            \label{fig:workflow}
        \end{subfigure}
        \hfill
        \begin{subfigure}[b]{0.3\textwidth}
            \centering
            \includegraphics[width=\textwidth,trim=8pt 8pt 8pt 8pt,clip]{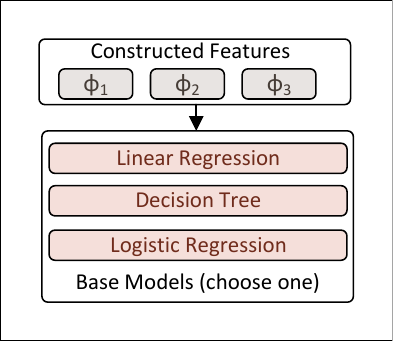}
            \caption{Evaluation.}
            \label{fig:evaluation}
        \end{subfigure}
        \caption{(a) overall workflow from population initialization through elitism. (b) evaluation of the evolutionary feature construction process.}
        \label{fig:workflow-evaluation}
        \vspace{-4mm}
    \end{figure}

    The evolutionary feature construction process follows a generational evolutionary algorithm framework~\cite{banzhaf1998genetic}. Let $\Phi = \{\Phi_1, \Phi_2, \ldots, \Phi_N\}$ denote the population of $N$ individuals, where each individual $\Phi_i$ is a set of feature transformation trees $\{\phi_1^{(i)}, \phi_2^{(i)}, \ldots, \phi_k^{(i)}\}$. Each tree $\phi_j^{(i)}$ is constructed using the primitives of GP, which include mathematical operators and terminal nodes representing input variables or constants. \Cref{fig:workflow} illustrates the overall workflow of the evolutionary feature construction process. The workflow consists of seven main stages:

    \begin{itemize}
        \setlength{\itemsep}{0pt}\setlength{\parskip}{0pt}
        \item \textbf{Population Initialization:} An initial population $\Phi^{(0)}$ of $N$ individuals is created, where each individual $\Phi_i^{(0)}$ contains multiple trees $\{\phi_1^{(i)}, \ldots, \phi_k^{(i)}\}$, and each tree $\phi_j^{(i)}$ is randomly generated using the ramped half-and-half initialization method~\cite{banzhaf1998genetic} with GP primitives.

        \item \textbf{Solution Evaluation:} For each individual $\Phi_i$ in the current population, a machine learning model is trained on the features constructed from its trees $\{\phi_1^{(i)}, \phi_2^{(i)}, \ldots, \phi_k^{(i)}\}$. The fitness $f(\Phi_i)$ is a scalar measuring prediction performance: coefficient of determination ($R^2$) in regression and area under the ROC curve (AUC) in classification. The loss vector has one component per training example and is used by epsilon lexicase selection for parent selection: per-example squared error in regression and per-example cross-entropy in classification. For regression tasks, we consider ridge regression and random decision trees. For classification tasks, we use logistic regression:
        \begin{itemize}
            \setlength{\itemsep}{0pt}\setlength{\parskip}{0pt}
            \item \textbf{Ridge Regression~\cite{la2020learning}:} The method uses ridge regression with efficient leave-one-out evaluation for model training and fitness computation.
            \item \textbf{Random Decision Tree~\cite{zhang2021evolutionary}:} This method uses a decision tree with a random splitter, randomly selecting splits from candidates rather than choosing optimal splits. Random decision trees are chosen because they have been shown to perform better than standard decision trees in evolutionary feature construction~\cite{zhang2021evolutionary}. Five-fold cross-validation error is used as fitness.
            \item \textbf{Logistic Regression:} The base learner is regularized logistic regression, mapping the constructed feature vector to class probabilities:
            \begin{equation}
                \mathbb{P}(y = 1 \mid \mathbf{x}; \boldsymbol{\beta}) = \sigma\bigl(\beta_0 + \boldsymbol{\beta}^\top \boldsymbol{\phi}(\mathbf{x})\bigr) = \sigma\biggl(\beta_0 + \sum_{j=1}^{k} \beta_j \,\phi_j^{(i)}(\mathbf{x})\biggr),
            \end{equation}
            where $\sigma(z) = 1/(1 + e^{-z})$ is the logistic function and $\boldsymbol{\beta} = (\beta_1, \ldots, \beta_k)^\top$ are the coefficients for the constructed features. Similarly, five-fold cross-validation score is used as fitness.
        \end{itemize}

        \item \textbf{Importance Calculation:} After evaluation, feature importance $I_j^{(i)}$ is calculated for each tree $\phi_j^{(i)}$ using the importance method for the chosen base learner, as detailed in \Cref{sec:feature-importance}. For ridge regression, we use one of three options: built-in coefficients, SHAP values, or marginal contribution. For random decision trees, we use impurity-decrease importance averaged over the five cross-validation folds. For logistic regression in credit classification, we use the absolute coefficient $|\beta_j|$. These importance values are stored with each individual and form the basis of the protection mechanism.

        \item \textbf{Parent Selection:} Parent individuals $\Phi_p$ and $\Phi_q$ are selected from the current population $\Phi$ using epsilon lexicase selection~\cite{helmuth2014solving}, where $p, q \in \{1, 2, \ldots, N\}$. The $\epsilon$ threshold is set to the median absolute deviation (MAD) of per-example loss values~\cite{la2019probabilistic}, with the selection threshold given by $\min + \epsilon$, where $\min$ is the minimum per-example loss value over the population.

        \item \textbf{Offspring Generation:} New individuals $\Phi_{\text{offspring}}$ are created through genetic operations applied to selected parent individuals. Random subtree crossover combines trees from two parents $\Phi_p$ and $\Phi_q$ to create offspring by exchanging randomly selected subtrees~\cite{banzhaf1998genetic}, while random subtree mutation introduces random modifications to individual trees $\phi_j$ by replacing randomly selected subtrees. In multi-tree GP, where $k$ is the number of trees in each individual, these crossover and mutation operators can be invoked $k$ times with their corresponding probabilities, allowing multiple trees within an individual to be modified during offspring generation.

        \item \textbf{Adaptive Protection:} The adaptive protection mechanism, detailed in \Cref{sec:adaptive-protection}, operates in two phases. Before genetic operations, the original trees $\phi_j^{\text{original}}$ are saved. After crossover and mutation, selected trees are restored to this saved state using an importance-weighted protection score.

        \item \textbf{Elitism:} Two mechanisms are employed to preserve valuable genetic material:
        \begin{itemize}
            \setlength{\itemsep}{0pt}\setlength{\parskip}{0pt}
            \item \textbf{Traditional Elitism:} The best-performing individual from the current generation is carried forward to the next generation to preserve valuable genetic material.
            \item \textbf{Ensemble Archive:} To improve prediction stability, we maintain a separate archive of the top 100 evaluated individuals and use it for final prediction. When the base learner is a random decision tree, predictions from the archived individuals are averaged to reduce variance caused by the stochastic tree-construction process. In credit classification, where the base learner is logistic regression, the same strategy averages predicted class probabilities from the archived individuals, which helps reduce sensitivity to noise in the datasets. The archive is used for prediction, not for generating new solutions. We do not use this archive with ridge regression because ridge is deterministic and already stable.
        \end{itemize}
    \end{itemize}

    \subsection{Feature Importance Calculation}\label{sec:feature-importance}

    Feature importance serves as the foundation for the protection mechanism. The framework is general and can be applied with different types of importance metrics. To demonstrate this generality, we evaluate three calculation methods for ridge regression---namely built-in coefficients, SHAP values, and marginal contribution---each measuring feature importance from a different perspective. For random decision trees, we use the built-in method based on impurity decrease. For logistic regression in credit classification, we use the absolute coefficient as the built-in importance measure. The selected method is applied consistently throughout evolution for all individuals in the population at every generation, and in the adaptive protection mechanism the importance $I_j^{(i)}$ of each constructed feature $\phi_j^{(i)}$ is computed accordingly. A primary objective of this work is to systematically compare the effectiveness of the three feature importance metrics when used within the protection mechanism.

    \subsubsection{Built-In Coefficients}

    The built-in method uses model-specific importance metrics directly, providing computational efficiency. Two variants are used depending on the base learner:

    \begin{itemize}
        \setlength{\itemsep}{0pt}\setlength{\parskip}{0pt}
        \item \textbf{Ridge Regression and Logistic Regression:} For linear models such as ridge regression and logistic regression, the importance of constructed feature $j$ is computed as the absolute value of its coefficient: $I_j^{\text{built-in}} = |\beta_j|$, where $\beta_j$ is the coefficient for the constructed feature $j$ in the model. The constructed features are standardized before fitting, ensuring $|\beta_j|$ reflects feature contribution rather than scale. This method offers minimal computational overhead and provides a direct measure of feature contribution as encoded by the model.

        \item \textbf{Random Decision Tree:} For decision tree base learners, the importance of feature $j$ is computed by summing the weighted impurity decrease across all splits that use feature $j$: $I_j^{\text{built-in}} = \sum_{s \in S_j} w_s \cdot \Delta I_s$, where $S_j$ is the set of all splits that use feature $j$, $w_s$ is the fraction of samples reaching split $s$, and $\Delta I_s$ is the impurity decrease at split $s$, defined as $\Delta I_s = I_{\text{parent}} - \bigl( \frac{n_{\text{left}}}{n_s} I_{\text{left}} + \frac{n_{\text{right}}}{n_s} I_{\text{right}} \bigr)$. Here, $I_{\text{parent}}$, $I_{\text{left}}$, and $I_{\text{right}}$ are the impurities of the parent node and left/right child nodes, respectively, measured as variance for regression, and $n_s$, $n_{\text{left}}$, and $n_{\text{right}}$ are the numbers of samples at the parent and child nodes. The final importance values are normalized to sum to 1.
    \end{itemize}

    \subsubsection{SHAP Values}

    Shapley values offer a game-theoretic approach to feature importance~\cite{lundberg2017unified}, computing the average marginal contribution of each feature across all possible feature subsets. The SHAP value for feature $j$ is defined as: $\psi_j = \sum_{S \subseteq F \setminus \{j\}} \frac{|S|!(|F| - |S| - 1)!}{|F|!} [v(S \cup \{j\}) - v(S)]$, where $F$ is the set of all features, $S$ is a subset of features excluding $j$, and $v(S)$ is the value function representing the model's prediction when only features in $S$ are used. For linear models, LinearSHAP~\cite{lundberg2017unified} gives a closed-form per-instance SHAP value $\beta_j(x_{ij}-\mathbb{E}[X_j])$. Per-instance SHAP values can be positive or negative. We define SHAP importance $I_j^{\text{SHAP}}$ for feature $j$ as the mean absolute SHAP value over the $n$ evaluation instances:
    \begin{equation}
        I_j^{\text{SHAP}} = \frac{1}{n}\sum_{i=1}^{n} \bigl|\beta_j(x_{ij}-\mathbb{E}[X_j])\bigr|.
    \end{equation}
    Here $\beta_j$ is the coefficient for feature $j$, $x_{ij}$ is the value of feature $j$ for instance $i$, and $\mathbb{E}[X_j]$ is the expected value of feature $j$. Because $I_j^{\text{SHAP}}$ averages $|\beta_j(x_{ij}-\mathbb{E}[X_j])|$ over instances, this scalar importance reflects both coefficient magnitude and the distribution of feature $j$, not coefficient magnitude alone.

    \subsubsection{Marginal Contribution}

    In this paper, we define marginal contribution as how much a constructed feature contributes to predictive performance, measured by comparing the $R^2$ of two ridge-regression models~\cite{behzadifar2025decompose}. The full model is the ridge-regression model fitted using the complete set of constructed features in the individual, whereas the reduced model for feature $j$ is obtained by removing feature $j$ from that set and refitting the ridge-regression model on the remaining constructed features. Let $R^2_{\text{full}}$ denote the score of the full model and $R^2_{-j}$ the score of the reduced model for feature $j$. The marginal contribution of feature $j$ is $I_j^{\text{MC}} = R^2_{\text{full}} - R^2_{-j}$. The definition is intuitive and model-agnostic, and unlike fixed-model measures it explicitly accounts for refitting. The trade-off is one additional model fit per constructed feature, so it is more expensive than built-in coefficient importance. When constructed features are correlated, removing one may be offset by others, so marginal contribution can underestimate importance for multiple correlated features and leave them more exposed to genetic operators, with possible performance loss. We include this method to quantify these effects and to examine how the choice of importance metric affects the protection mechanism.

    \subsection{Adaptive Protection Mechanism}\label{sec:adaptive-protection}

    \begin{algorithm}[!t]
    \footnotesize
    \caption{Adaptive Protection Mechanism (APM)}
    \label{alg:protection}
    \begin{algorithmic}[1]
        \Require Individual $\Phi_i$ with trees $\{\phi_1^{(i)}, \phi_2^{(i)}, \ldots, \phi_k^{(i)}\}$ and importance values $\{I_1^{(i)}, I_2^{(i)}, \ldots, I_k^{(i)}\}$, protection coefficient $m_{\text{protect}}$
        \Ensure Modified individual $\Phi_i'$ after genetic operations with selective protection
        \State Save original trees: $\phi_j^{\text{original}} \gets \phi_j^{(i)}$ for all $j \in \{1, 2, \ldots, k\}$ \label{alg:protection:save}
        \State $\phi_j^{(i)} \gets \phi_j^{\text{modified}} \gets \text{Crossover/Mutation}(\phi_j^{(i)})$ for all $j \in \{1, \ldots, k\}$ \label{alg:protection:apply}
        \State Normalize importance values: $\tilde{I}_j^{(i)} \gets |I_j^{(i)}| / \sum_{j=1}^{k} |I_j^{(i)}|$ for all $j \in \{1, 2, \ldots, k\}$ \label{alg:protection:normalize}
        \For{$j = 1$ to $k$}
            \label{alg:protection:loop-start}
            \State Sample $u \sim \text{Uniform}(0, 1)$
            \State $s_j \gets m_{\text{protect}} \cdot \tilde{I}_j^{(i)}$ \Comment{Protection score}
            \If{$u < s_j$}
                \State $\phi_j^{(i)} \gets \phi_j^{\text{original}}$ \Comment{Restore to protect important features}
            \EndIf
        \EndFor \label{alg:protection:loop-end}
        \State \Return $\Phi_i'$ with updated trees
    \end{algorithmic}
    \end{algorithm}

    As shown in Algorithm~\ref{alg:protection}, the mechanism saves the original trees, applies genetic operations, then computes a protection score $s_j = m_{\text{protect}} \cdot \tilde{I}_j^{(i)}$ for each tree, where $\tilde{I}_j^{(i)}$ is the normalized importance. Each tree is restored to its original state when a uniform random draw falls below this score. Thus, higher-importance trees are restored more often, reducing the chance that useful building blocks are lost, while lower-importance trees are still frequently modified or replaced, allowing the search to continue generating and testing new feature combinations. The protection coefficient $m_{\text{protect}}$ controls protection strength. A value of zero disables protection. Larger values provide stronger protection but may reduce diversity. The saved trees are the same as those already stored for depth-limit checking, avoiding additional overhead. Conceptually, this can be viewed as a more adaptive form of elitism: rather than keeping entire individuals, it preserves the most valuable building blocks within each individual.

    \section{Experimental Setup}\label{sec:experimental-setup}

    \subsection{Datasets}

    The main experiments are conducted on 98 benchmark regression datasets from the Penn Machine Learning Benchmarks (PMLB) repository~\cite{olson2017pmlb}, focusing on datasets with up to 2000 samples due to computational cost constraints. To assess whether the proposed mechanism also applies beyond symbolic regression, we additionally evaluate it on two credit classification datasets: Australian Credit and German Credit~\cite{yu2022extreme}.

    \subsection{Parameter Settings}

    The experimental design compares different protection levels and feature importance calculation methods. Table~\ref{tab:parameters} summarizes the key parameter settings used in our experiments. The impact of the protection coefficient is further analyzed in \Cref{app:effective-protection-level}.\footnote{Supplementary material is available at \url{https://github.com/hengzhe-zhang/ppsn2026-adaptive-protection/blob/main/supplementary_material.pdf}.} Categorical features are encoded using target encoding~\cite{micci2001preprocessing} fitted on the training data. The function set includes arithmetic operations, mathematical functions, and the analytical quotient $\mathrm{AQ}(x, y) = x / \sqrt{1 + y^2}$. For the credit classification experiments, the number of generations is reduced to 30 to avoid overfitting, while all other settings remain the same.

    \begin{table}[!t]
        \centering
        \footnotesize
        \caption{Parameter Settings}
        \label{tab:parameters}
        \adjustbox{scale=0.85,center}
        {
            \begin{tabular}{lc}
                \toprule
                \textbf{Parameter}             & \textbf{Value}                                                                                               \\
                \midrule
                Population Size                & 200                                                                                                          \\
                Number of Generations          & 100                                                                                                          \\
                Crossover Probability          & 0.9                                                                                                          \\
                Mutation Probability           & 0.1                                                                                                          \\
                Protection Coefficient         & 5                                                                                                            \\
                Max Initial Tree Depth         & 2                                                                                                             \\
                Maximum Tree Depth             & 10                                                                                                           \\
                Number of Trees per Individual & 10                                                                                                           \\
                Elitism, Number of Individuals & 1                                                                                                            \\
                Functions                      & \makecell[l]{$+,\, -,\, \times,\, \mathrm{AQ},\, \sqrt{|\cdot|},\, \log(1+|\cdot|),\, |\cdot|,\, (\cdot)^2,$ \\
                    $\sin_\pi(\cdot),\, \cos_\pi(\cdot),\, \mathrm{Max},\, \mathrm{Min},\, \mathrm{Neg}$}                                                          \\
                \bottomrule
            \end{tabular}
        }
        \vspace{-4mm}
    \end{table}

    \subsection{Evaluation Protocol}

    Performance is measured using R² score on test sets for regression experiments, which is scale-insensitive, and using area under the ROC curve (AUC) for the credit classification experiments, which is appropriate for imbalanced datasets. Each experiment is conducted with 30 independent runs, using an 80:20 train-test split. Results are averaged across multiple runs, and statistical significance is tested using the Wilcoxon signed-rank test with $p < 0.05$. Research questions RQ1--RQ3 are addressed using ridge regression as the base learner, RQ4 uses a random decision tree as the base learner to evaluate robustness across different base learners, and RQ5 evaluates the application to credit classification using a class-balanced, regularized logistic regression base learner and coefficient-based feature importance.

    \section{Results and Analysis}\label{sec:results}

    We investigate the following research questions:
    \begin{itemize}
        \setlength{\itemsep}{0pt}\setlength{\parskip}{0pt}
        \item \textbf{RQ1:} Does protection improve predictive performance compared with no protection?
        \item \textbf{RQ2:} Is adaptive protection more effective than reducing genetic operations?
        \item \textbf{RQ3:} What feature importance method provides the best balance between effectiveness and efficiency?
        \item \textbf{RQ4:} Is the proposed method robust when applied with different base learners?
        \item \textbf{RQ5:} Does the adaptive protection mechanism transfer effectively to credit classification?
    \end{itemize}

    \begin{table}[!t]
        \centering
        \footnotesize
        \caption{Statistical comparison of adaptive protection strategies and baselines on \textbf{training R² score}. Each cell reports the number of datasets with wins/ties/losses, denoted by $+$, $\sim$, and $-$ respectively, for the row method against the column method.}
        \label{tab:importance_methods_comparison_train}
        \resizebox{\textwidth}{!}{%
            \begin{tabular}{ccccccc}
                \toprule

                &\textbf{No Protect}&\textbf{Low Cross/Mut (1)}&\textbf{Low Cross/Mut (5)}&\textbf{APM (SHAP)}&\textbf{APM (Marg. Contrib.)}\\%
                \midrule%
                \textbf{APM (Built-In)}&93(+)/5($\sim$)/0({-})&78(+)/20($\sim$)/0({-})&89(+)/9($\sim$)/0({-})&4(+)/93($\sim$)/1({-})&9(+)/77($\sim$)/12({-})\\%
                \textbf{No Protect}&---&4(+)/12($\sim$)/82({-})&0(+)/11($\sim$)/87({-})&0(+)/5($\sim$)/93({-})&0(+)/3($\sim$)/95({-})\\%
                \textbf{Low Cross/Mut (1)}&---&---&29(+)/60($\sim$)/9({-})&1(+)/19($\sim$)/78({-})&0(+)/19($\sim$)/79({-})\\%
                \textbf{Low Cross/Mut (5)}&---&---&---&1(+)/9($\sim$)/88({-})&0(+)/9($\sim$)/89({-})\\%
                \textbf{APM (SHAP)}&---&---&---&---&7(+)/76($\sim$)/15({-})\\%
                \bottomrule
            \end{tabular}%
        }
        \vspace{-4mm}
    \end{table}

    \begin{table}[!t]
        \centering
        \footnotesize
        \caption{Statistical comparison of adaptive protection strategies and baselines on \textbf{test R² score}. Each cell reports the number of datasets with wins/ties/losses, denoted by $+$, $\sim$, and $-$ respectively, for the row method against the column method.}
        \label{tab:importance_methods_comparison}
        \resizebox{\textwidth}{!}{%
            \begin{tabular}{ccccccc}
                \toprule

                &\textbf{No Protect}&\textbf{Low Cross/Mut (1)}&\textbf{Low Cross/Mut (5)}&\textbf{APM (SHAP)}&\textbf{APM (Marg. Contrib.)}\\%
                \midrule%
                \textbf{APM (Built-In)}&56(+)/39($\sim$)/3({-})&34(+)/61($\sim$)/3({-})&47(+)/49($\sim$)/2({-})&1(+)/93($\sim$)/4({-})&5(+)/88($\sim$)/5({-})\\%
                \textbf{No Protect}&---&0(+)/44($\sim$)/54({-})&0(+)/48($\sim$)/50({-})&1(+)/37($\sim$)/60({-})&6(+)/34($\sim$)/58({-})\\%
                \textbf{Low Cross/Mut (1)}&---&---&14(+)/82($\sim$)/2({-})&1(+)/68($\sim$)/29({-})&5(+)/52($\sim$)/41({-})\\%
                \textbf{Low Cross/Mut (5)}&---&---&---&5(+)/46($\sim$)/47({-})&6(+)/46($\sim$)/46({-})\\%
                \textbf{APM (SHAP)}&---&---&---&---&5(+)/92($\sim$)/1({-})\\%
                \bottomrule
            \end{tabular}%
        }
        \vspace{-4mm}
    \end{table}

    \begin{table}[!t]
        \centering
        \footnotesize
        \caption{Comparison of constructed features before and after variation with and without adaptive protection on OpenML 586.}
        \label{tab:protection_comparison}
        \adjustbox{max width=\linewidth}{%
            \begin{tabular}{r l r l r l}
                \toprule                 Coef (Orig) & Expr (Orig)                                 & Coef (No Prot) & Expr (No Prot)                              & Coef (Prot) & Expr (Prot)                                 \\
                \midrule
                2.423               & AQ($X_{0}$, $X_{1}$)                        & 0.02242        & AQ(Add($X_{1}$, $X_{9}$), $X_{1}$)          & 2.621       & AQ($X_{0}$, $X_{1}$)                        \\
                1.802               & Add($X_{0}$, AbsLog(Max($X_{3}$, $X_{4}$))) & 1.096          & Add($X_{0}$, AbsLog(Max($X_{0}$, $X_{4}$))) & 1.818 & Add($X_{0}$, AbsLog(Max($X_{3}$, $X_{4}$))) \\
                1.196               & $X_{0}$                                     & 0.4186         & $X_{0}$                                     & 1.415       & $X_{0}$                                     \\
                0.8335              & $X_{1}$                                     & 0.363          & $X_{1}$                                     & 0.8738      & $X_{1}$                                     \\
                0.4052              & CosPi(Max($X_{1}$, $X_{0}$))                & 0.2998         & CosPi(Max($X_{1}$, $X_{13}$))               & 0.3173      & CosPi(Max($X_{1}$, $X_{0}$))                \\
                0.187               & Abs($X_{3}$)                                & 0.1201         & Abs($X_{0}$)                                & 0.2778      & Abs($X_{3}$)                                \\
                0.03374             & $X_{21}$                                    & 0.01486        & $X_{21}$                                    & 0.1226      & CosPi($X_{1}$)                              \\
                0.01167             & Square($X_{13}$)                            & 0.4186         & $X_{0}$                                     & 0.08625     & Square($X_{3}$)                             \\
                0.008119            & Abs($X_{0}$)                                & 0.003573       & Sqrt(Abs($X_{21}$))                         & 0.0003979   & Abs($X_{0}$)                                \\
                0.001442            & AbsLog(Neg($X_{23}$))                       & 0.01326        & AbsLog(Neg($X_{23}$))                       & 0.0247      & AbsLog(Neg(SinPi($X_{1}$)))                 \\
                \bottomrule
            \end{tabular}%
        }
        \vspace{-4mm}
    \end{table}

    \subsection{RQ1: Protection vs.\ No Protection}

    \begin{figure}[!t]
        \centering
        \begin{subfigure}[b]{0.48\textwidth}
            \centering
            \includegraphics[width=\textwidth]{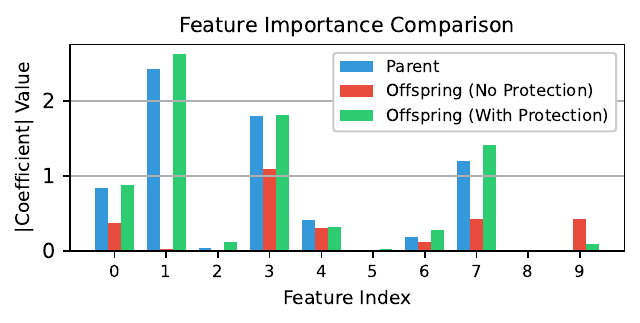}
            \caption{Impact of Adaptive Protection on Feature Importance}
            \label{fig:protection_importance}
        \end{subfigure}
        \hfill
        \begin{subfigure}[b]{0.4\textwidth}
            \centering
            \includegraphics[width=\textwidth]{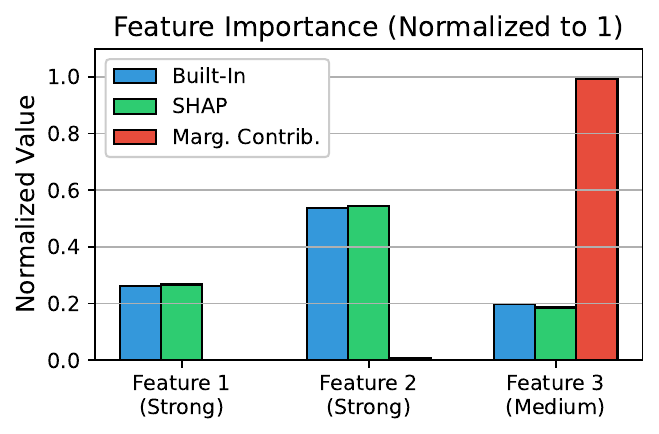}
            \caption{Marginal Contribution Pitfall}
            \label{fig:partial_r2_pitfall}
        \end{subfigure}
        \caption{The left panel shows the impact of adaptive protection on feature importance on OpenML 586. The right panel shows that marginal contribution can fail to identify truly useful features when highly correlated features coexist and can compensate for one another, illustrated on a synthetic dataset.}
        \vspace{-4mm}
    \end{figure}

    \newcommand{\builtinwins}{56}
    \newcommand{\builtinlosses}{3}
    \newcommand{\shapwins}{60}
    \newcommand{\shaplosses}{one}
    \newcommand{\partialrsquaredwins}{58}
    \newcommand{\partialrsquaredlosses}{6}
    \newcommand{\totaldatasets}{98}
    \newcommand{\parenttestr}{0.800}
    \newcommand{\withprotectiontestr}{0.814}
    \newcommand{\withoutprotectiontestr}{0.101}

    To evaluate whether adaptive protection improves solution quality, we compare configurations with protection, where the protection level is greater than zero, against the baseline multi-tree genetic programming (MTGP) configuration without protection. \Cref{tab:importance_methods_comparison} presents the test-set statistical comparison across different configurations, and \Cref{tab:importance_methods_comparison_train} presents the corresponding training-set comparison.

    The results demonstrate that protection mechanisms consistently outperform the no-protection baseline across different importance calculation methods. As shown in \Cref{tab:importance_methods_comparison}, Built-In wins on \builtinwins{} of \totaldatasets{} datasets with only \builtinlosses{} losses, SHAP wins on \shapwins{} datasets with only \shaplosses{} loss, and Marginal Contribution wins on \partialrsquaredwins{} datasets with only \partialrsquaredlosses{} losses. This pattern holds consistently regardless of the importance calculation method used, indicating robustness to the choice of importance metric. \Cref{fig:protection_importance} illustrates this effect on the OpenML 586 dataset: without protection, the feature-importance values after variation differ substantially from the original values, whereas with adaptive protection they remain much closer. This suggests that, without protection, genetic variation changes the solution too aggressively and leads to worse search locality, meaning that offspring are less likely to preserve the behavior of their parents. \Cref{tab:protection_comparison} further illustrates this effect by showing the coefficients and symbolic expressions of constructed features before and after genetic variation, with and without adaptive protection. For example, the most important feature $\mathrm{AQ}(X_{0}, X_{1})$ is disrupted without protection: its expression changes to $\mathrm{AQ}(\mathrm{Add}(X_{1}, X_{9}), X_{1})$, whereas with protection the original expression is preserved. For the individual in \Cref{tab:protection_comparison}, the parent test R² is \parenttestr{}, which drops to \withoutprotectiontestr{} without protection, indicating severe degradation in solution quality. However, with adaptive protection, the test R² improves to \withprotectiontestr{}, demonstrating that the protection mechanism successfully maintains solution quality during genetic operations.

    \begin{figure}[!t]
        \centering
        \begin{subfigure}[b]{0.48\textwidth}
            \centering
            \includegraphics[width=\textwidth]{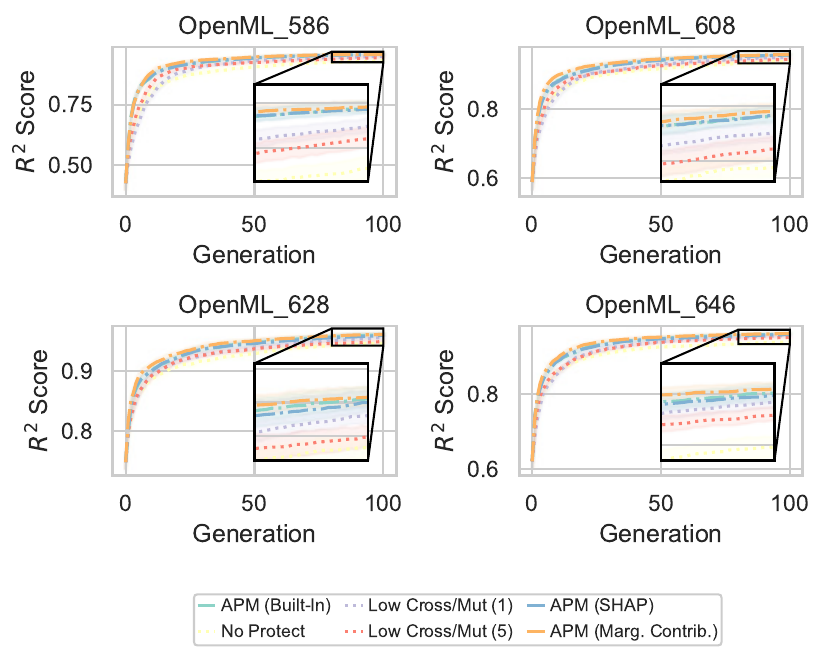}
            \caption{Training R² Convergence Curves}
            \label{fig:importance_train_r2}
        \end{subfigure}
        \hfill
        \begin{subfigure}[b]{0.48\textwidth}
            \centering
            \includegraphics[width=\textwidth]{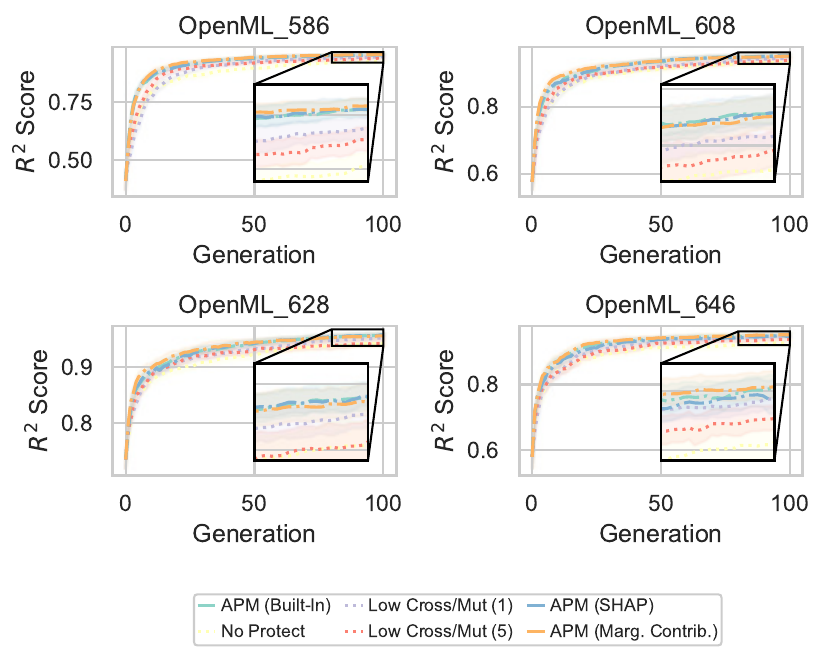}
            \caption{Test R² Convergence Curves}
            \label{fig:importance_test_r2}
        \end{subfigure}
        \caption{Convergence curves of different feature importance methods on four representative datasets.}
        \label{fig:importance_r2_comparison}
        \vspace{-4mm}
    \end{figure}

    \subsection{RQ2: Adaptive Protection vs. Reduced Genetic Operations}

    A key question is whether the observed improvements stem from applying fewer genetic operations or from the adaptive protection mechanism itself. To address this, we compare standard operation rates against reduced rates, where ``Low Cross/Mut (1)'' and ``Low Cross/Mut (5)'' in \Cref{tab:importance_methods_comparison} represent restricted-variation baselines in which genetic operators are applied to only 1 or 5 trees per offspring with corresponding crossover and mutation probabilities, rather than all 10 trees.

    The results reveal that simply reducing the number of genetic operations is less effective than adaptive protection. As shown in \Cref{tab:importance_methods_comparison}, all protection-based methods consistently outperform low operation rate configurations. This indicates that selective protection provides a more principled approach than globally reducing genetic operations, allowing broader modification of offspring while maintaining solution quality through selective preservation of valuable genetic material.
    \subsection{RQ3: Feature Importance Method Comparison}\label{subsec:rq3-importance-method-comparison}

    To understand which feature importance method offers the best balance between effectiveness and efficiency, we compare the training and test performance of three methods: SHAP values, marginal contribution, and Built-In coefficients, using \Cref{tab:importance_methods_comparison} for test results and \Cref{tab:importance_methods_comparison_train} for the corresponding training results. \Cref{fig:importance_r2_comparison} shows convergence curves of different feature importance methods on four representative datasets.

    The general conclusion is that most feature importance methods show remarkably similar performance, with the vast majority of comparisons between Built-In and SHAP resulting in ties. This pattern suggests that the choice of importance calculation method has a relatively modest impact on the overall effectiveness of the protection mechanism, as long as a reasonable importance metric is used.

    The comparison between Built-In and marginal contribution has mixed results. Built-In performs better when multicollinearity is present because marginal contribution can assign low scores to correlated features: removing one such feature may have little effect on $R^2$ after refitting because the remaining correlated feature can compensate for it. \Cref{fig:partial_r2_pitfall} illustrates this with an example in which features are defined as $f_1 = s + \epsilon_1$, $f_2 = s + \epsilon_2$, and $f_3$ is independent. In this setting, $s$ is a shared signal and $\epsilon_1, \epsilon_2$ are small independent noise terms. The target is $y = 5.0 \cdot f_1 + 5.0 \cdot f_2 + 2.5 \cdot f_3 + \epsilon$. The key point is that marginal contribution fails to identify the truly useful features: although $f_1$ and $f_2$ carry the stronger signal and have larger coefficients, they receive lower marginal contribution scores than the less important but uncorrelated feature $f_3$. This becomes problematic when the protection mechanism relies on these scores, because the features that should receive stronger protection may instead be treated as unimportant. In other cases, however, marginal contribution can perform better because it more directly reflects a feature's contribution to predictive performance.

\begin{figure}[!t]
        \centering
        \begin{subfigure}[b]{0.48\textwidth}
            \centering
            \includegraphics[width=0.75\textwidth]{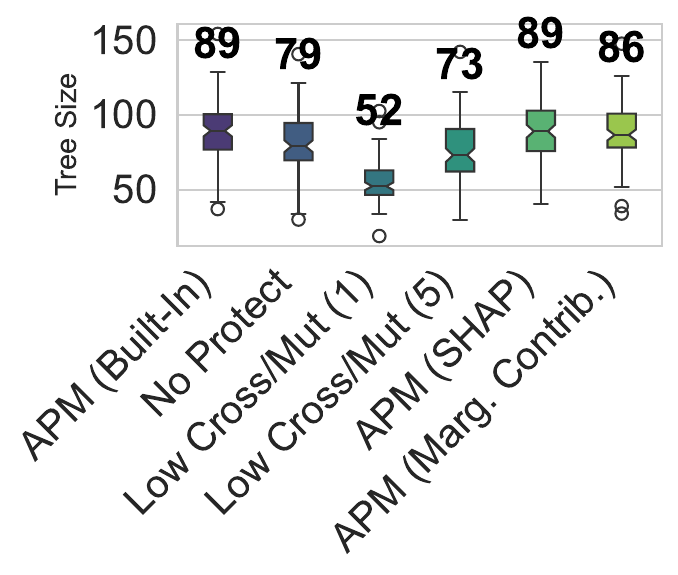}
            \caption{Tree Size}
            \label{fig:importance_complexity}
        \end{subfigure}
        \hfill
        \begin{subfigure}[b]{0.48\textwidth}
            \centering
            \includegraphics[width=0.75\textwidth]{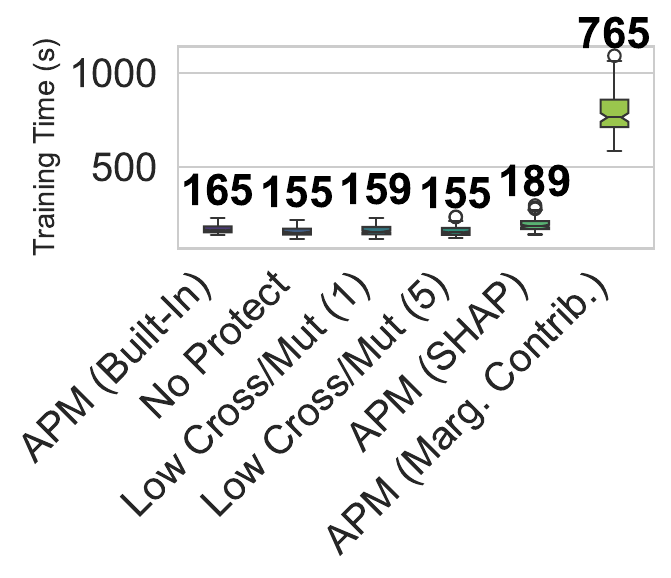}
            \caption{Training Time}
            \label{fig:importance_time}
        \end{subfigure}
        \caption{Comparison of feature importance methods in terms of complexity and training time.}
        \label{fig:importance_rq3}
        \vspace{-4mm}
    \end{figure}

    From a computational efficiency perspective, Built-In is the most efficient because it reuses fitted model coefficients with almost no extra overhead. Marginal contribution is computationally expensive because it requires one additional refit per feature for each evaluated individual. LinearSHAP is also more expensive than Built-In but typically cheaper than exhaustive refit-based importance. \Cref{fig:importance_rq3} shows the runtime impact of these choices. Given the broadly similar predictive effectiveness of these methods, Built-In remains attractive in practice because of its low cost.

    \subsection{RQ4: Robustness with Different Base Learners}

    To verify the robustness of the proposed method when used with other base learners in feature construction, we evaluate the adaptive protection mechanism using a random decision tree as the base learner. The main motivation is to verify that the mechanism remains effective when feature importance is computed from tree-based impurity decrease rather than linear model coefficients. Random decision trees use the impurity-decrease importance described in \Cref{sec:feature-importance}, which differs fundamentally from the coefficient-based importance used in ridge regression.

    \Cref{tab:ensemble_importance_methods_comparison} presents the statistical comparison when using a random decision tree as the base learner. Comparisons are made within the same base learner setting to isolate the protection mechanism's effect. The results demonstrate that adaptive protection maintains effectiveness when feature importance is calculated from impurity decrease. Protection-based configurations consistently outperform the no-protection baseline and reduced genetic operation rates, mirroring the findings for RQ2. This confirms robustness across different base learners and importance calculation paradigms.

    \begin{table}[!t]
        \centering
        \footnotesize
        \caption{Statistical comparison of different configurations with random decision tree base learners on \textbf{test R² score}.}
        \label{tab:ensemble_importance_methods_comparison}
        \adjustbox{max width=0.75\linewidth}{
            \begin{tabular}{cccc}
                \toprule

                &\textbf{No Protect}&\textbf{Low Cross/Mut (1)}&\textbf{Low Cross/Mut (5)}\\%
                \midrule%
                \textbf{APM (Built-In)}&52(+)/34($\sim$)/12({-})&39(+)/54($\sim$)/5({-})&36(+)/54($\sim$)/8({-})\\%
                \textbf{No Protect}&---&5(+)/50($\sim$)/43({-})&2(+)/45($\sim$)/51({-})\\%
                \textbf{Low Cross/Mut (1)}&---&---&5(+)/82($\sim$)/11({-})\\%
                \bottomrule
            \end{tabular}
        }
        \vspace{-4mm}
    \end{table}

    \subsection{RQ5: Application to Credit Classification}\label{subsec:rq5-credit-classification}

    To verify that the adaptive protection mechanism transfers beyond symbolic regression, we evaluate it on the Australian and German credit datasets. \Cref{tab:classification_comparison_train} and \Cref{tab:classification_comparison_test} report training and test AUC. On training data, the proposed method achieves higher AUC than both no protection and reduced genetic operations on both datasets, showing that adaptive protection improves search effectiveness. On test data, it matches no protection on both datasets and outperforms reduced genetic operations on Australian while matching them on German. The equal test results suggest that protection offers limited benefit on unseen data here, possibly because the current function set is not designed for credit tasks, and a more tailored function set may be needed to reveal clearer gains.

    \begin{table}[!t]
        \centering
        \footnotesize
        \begin{minipage}[t]{0.48\textwidth}
            \centering
            \caption{Detailed \textbf{training AUC} for classification. $+$/$-$/$=$: better/worse/equal compared to Built-In.}
            \label{tab:classification_comparison_train}
            \resizebox{\linewidth}{!}{%
                \begin{tabular}{cccc}
                    \toprule
                    \textbf{}           & \textbf{Built-In} & \textbf{No Protect} & \textbf{Low Cross/Mut (1)} \\
                    \midrule
                    \textbf{Australian} & 0.961             & 0.956  (-)            & 0.957  (-)                   \\
                    \textbf{German}     & 0.844             & 0.832  (-)            & 0.834  (-)                   \\
                    \bottomrule
                \end{tabular}%
            }
        \end{minipage}
        \hfill
        \begin{minipage}[t]{0.48\textwidth}
            \centering
            \caption{Detailed \textbf{test AUC} for classification. $+$/$-$/$=$: better/worse/equal compared to Built-In.}
            \label{tab:classification_comparison_test}
            \resizebox{\linewidth}{!}{%
                \begin{tabular}{cccc}
                    \toprule
                    \textbf{}           & \textbf{Built-In} & \textbf{No Protect} & \textbf{Low Cross/Mut (1)} \\
                    \midrule
                    \textbf{Australian} & 0.942             & 0.941  (=)            & 0.934  (-)                   \\
                    \textbf{German}     & 0.787             & 0.787  (=)            & 0.787  (=)                   \\
                    \bottomrule
                \end{tabular}%
            }
        \end{minipage}
        \vspace{-4mm}
    \end{table}

    \section{Conclusions}\label{sec:conclusion}

    This paper introduced an adaptive protection mechanism for evolutionary feature construction that uses feature importance to guide the preservation of valuable genetic material. The mechanism restores selected constructed features to their original state after genetic operations, using an importance-weighted protection score, preserving important features while still allowing less important features to be modified or replaced.

    Experimental evaluation on 98 regression benchmark datasets demonstrates that protection-based methods substantially outperform no-protection baselines, and adaptive protection proves more effective than simply reducing genetic operations. Experiments on two credit classification datasets show that the mechanism also extends effectively to improve search effectiveness beyond symbolic regression. Across these settings, the mechanism shows robustness to different feature importance calculation methods and base learners, confirming its broad applicability to evolutionary feature construction.

    One potential limitation is that the protection mechanism operates at the feature level. A promising direction is to dynamically identify and protect useful substructures using building-block analysis techniques throughout the evolutionary process.

    \section*{Acknowledgements}
    This work was supported in part by the Marsden Fund of New Zealand Government under Contract VUW1913, and Contract VUW2016; in part by the MBIE Data Science SSIF Fund under Contract RTVU1914; in part by the MBIE Endeavor Research Programme under Contract UOCX2104; and in part by the Catalyst: Leaders International Leader Fellowship grant under Contract 23-VUW-006-ILF.

    \bibliographystyle{splncs04}
    \bibliography{mybibliography}

    \clearpage
    \appendix

    This appendix provides supplementary material supporting the empirical analysis in \Cref{sec:results}.

    \begin{itemize}
        \item \Cref{sec:appendix-credit-classification} provides additional background on why credit classification is a suitable application domain for evolutionary feature construction and supports the transferability discussion in \Cref{subsec:rq5-credit-classification}.
        \item \Cref{sec:appendix-protection-level} presents supplementary analysis of importance-value power transformations and effective protection levels, and supports the methodological comparison in \Cref{subsec:rq3-importance-method-comparison}.
        \item \Cref{sec:appendix-topk} compares the deterministic top-$K$ and probabilistic APM variants across different base learners.
    \end{itemize}

    \section{Feature Construction in Credit Classification}\label{sec:appendix-credit-classification}

    Credit classification is a suitable application for evolutionary feature construction because credit risk often depends on nonlinear interactions among attributes, and useful predictors may arise from complex combinations of raw features that are labor-intensive to design manually. Prior work shows that engineered transaction and behavioral features improve cost-sensitive fraud detection~\cite{bahnsen2016feature} and cash-out fraud ranking~\cite{wu2019feature}, while feature engineering with virtual sample generation improves robustness in data-scarce credit scoring~\cite{yu2022extreme}. A two-stage pipeline combining XGBoost-based feature construction with graph-based deep models also outperforms models trained on the original features~\cite{liu2022two}. However, these studies largely lack iterative refinement during search, motivating an evolutionary framework with adaptive protection to preserve high-value features while continuing the search for better feature combinations.

    \section{Further Analysis of Power Transformation and Protection Level}\label{sec:appendix-protection-level}\label{app:protection-level}

    \subsection{Power Transformation}

\begin{figure}[!ht]
        \centering
        \begin{subfigure}[b]{0.35\textwidth}
            \centering
            \includegraphics[width=\textwidth]{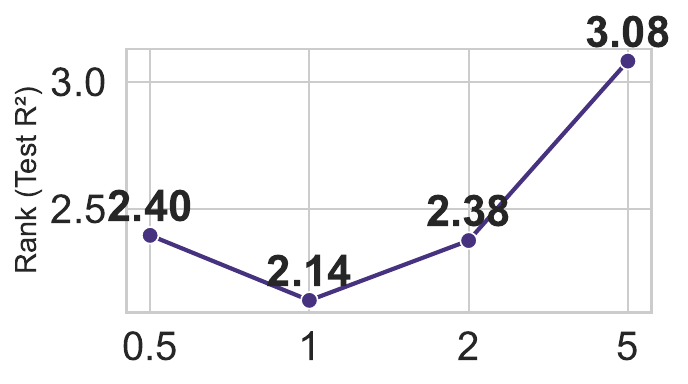}
            \caption{R² Ranks}
            \label{fig:power_r2}
        \end{subfigure}
        \hspace{0.04\textwidth}
        \begin{subfigure}[b]{0.35\textwidth}
            \centering
            \includegraphics[width=\textwidth]{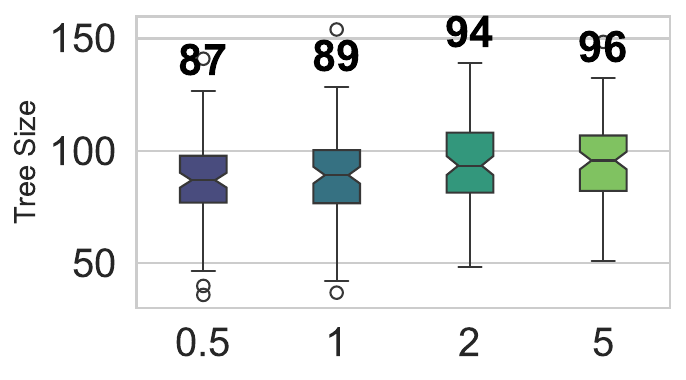}
            \caption{Tree Size}
            \label{fig:power_complexity}
        \end{subfigure}
        \caption{Comparison of feature importance power values.}
        \label{fig:power_transformation_comparison}
        \vspace{-4mm}
    \end{figure}

    As a complementary investigation within RQ3, we examine whether nonlinear transformations of importance values improve the protection mechanism by applying a power transformation $|\beta_j|^p$ for $p \in \{0.5, 1, 2, 5\}$. \Cref{fig:power_transformation_comparison} and \Cref{tab:power_comparison} show that power 1, corresponding to linear scaling, provides the best overall performance. Higher powers overemphasize differences, creating extreme rankings that underweight features of medium importance, leading to inadequate protection and performance degradation. Power 0.5 also underperforms relative to power 1, as it compresses importance values toward uniformity, reducing the contrast between high- and low-importance features and weakening the protection signal. Linear scaling therefore provides the most balanced approach, confirming that raw coefficient magnitudes are appropriate without requiring nonlinear transformations.

    \begin{table}[!ht]
        \centering
        \footnotesize
        \caption{Statistical comparison of feature importance power values.}
        \label{tab:power_comparison}
        \resizebox{0.48\textwidth}{!}{%
            \begin{tabular}{cccc}
                \toprule
                & \textbf{1}             & \textbf{2}             & \textbf{5}              \\%
                \midrule%
                \textbf{0.5} & 3(+)/92($\sim$)/3({-}) & 4(+)/91($\sim$)/3({-}) & 15(+)/81($\sim$)/2({-}) \\%
                \textbf{1}   & ---                    & 1(+)/97($\sim$)/0({-}) & 27(+)/70($\sim$)/1({-}) \\%
                \textbf{2}   & ---                    & ---                    & 20(+)/75($\sim$)/3({-}) \\%
                \bottomrule
            \end{tabular}%
        }
        \vspace{-4mm}
    \end{table}

    \subsection{Effective Protection Level}\label{app:effective-protection-level}

\begin{figure}[!ht]
        \centering
        \begin{subfigure}[b]{0.35\textwidth}
            \centering
            \includegraphics[width=\textwidth]{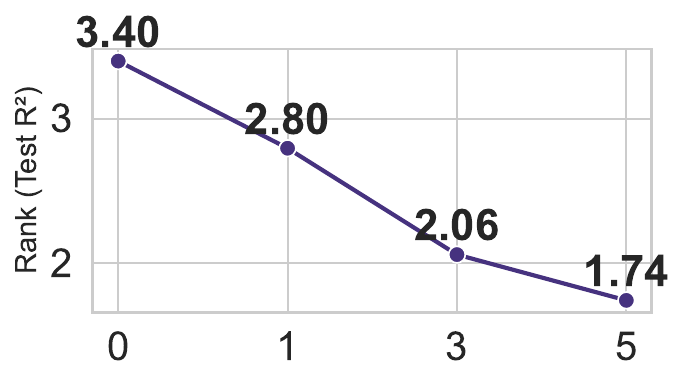}
            \caption{R² Ranks}
            \label{fig:protection_r2}
        \end{subfigure}
        \hspace{0.04\textwidth}
        \begin{subfigure}[b]{0.35\textwidth}
            \centering
            \includegraphics[width=\textwidth]{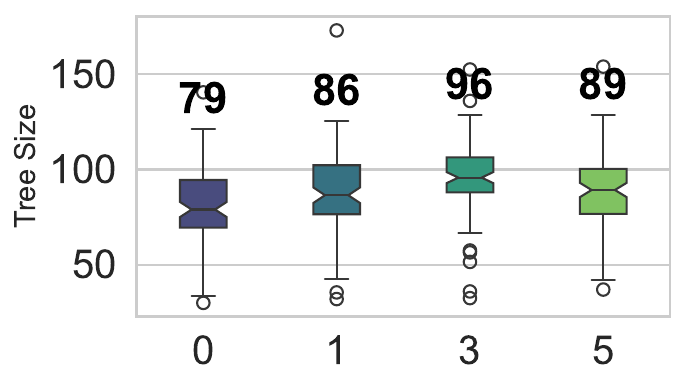}
            \caption{Tree Size}
            \label{fig:protection_complexity}
        \end{subfigure}
        \caption{Comparison of different protection coefficients.}
        \label{fig:protection_restoration_comparison}
        \vspace{-4mm}
    \end{figure}

    To determine an effective restoration intensity, we examine the trade-off between preserving existing valuable trees and allowing further tree modifications by comparing different protection coefficients. We consider protection coefficients of 0, 1, 3, and 5, where 0 represents no protection. \Cref{fig:protection_restoration_comparison} and \Cref{tab:protected_trees_comparison} summarize the outcomes. Statistical comparisons show that protection consistently outperforms no protection: with 10 trees per individual, larger protection coefficients perform better. Coefficients of 1 or 3 provide insufficient protection, whereas a coefficient of 5 provides adequate protection while still allowing sufficient modification of other trees.

    \begin{table}[!ht]
        \centering
        \footnotesize
        \caption{Statistical comparison of different protection coefficients.}
        \label{tab:protected_trees_comparison}
        \resizebox{0.48\textwidth}{!}{%
            \begin{tabular}{cccc}
                \toprule

                &\textbf{1}&\textbf{3}&\textbf{5}\\%
                \midrule%
                \textbf{0}&0(+)/67($\sim$)/31({-})&2(+)/44($\sim$)/52({-})&3(+)/39($\sim$)/56({-})\\%
                \textbf{1}&---&5(+)/51($\sim$)/42({-})&3(+)/42($\sim$)/53({-})\\%
                \textbf{3}&---&---&1(+)/75($\sim$)/22({-})\\%
                \bottomrule
            \end{tabular}%
        }
        \vspace{-4mm}
    \end{table}

    \section{Deterministic vs.\ Probabilistic Adaptive Protection}\label{sec:appendix-topk}

\begin{figure}[!ht]
        \centering
        \begin{subfigure}[b]{\textwidth}
            \centering
            \includegraphics[width=\textwidth]{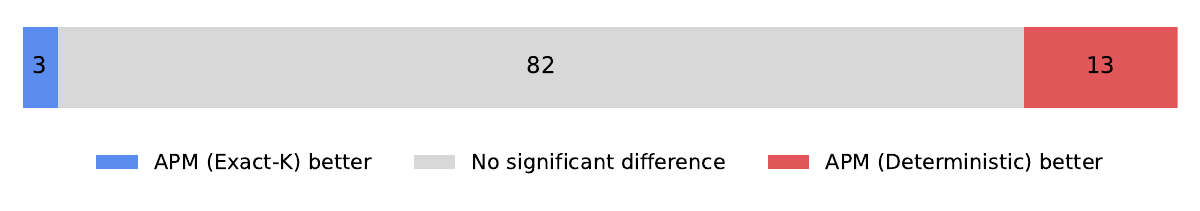}
            \caption{Ridge regression base learner.}
            \label{fig:topk_ridge}
        \end{subfigure}
        \begin{subfigure}[b]{\textwidth}
            \centering
            \includegraphics[width=\textwidth]{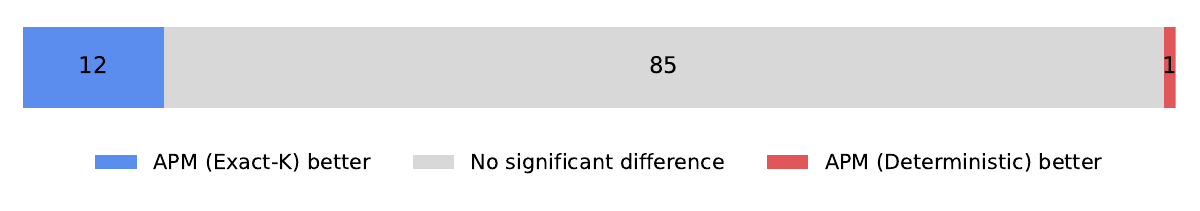}
            \caption{Random decision tree base learner.}
            \label{fig:topk_ensemble}
        \end{subfigure}
        \caption{Comparison of deterministic and probabilistic APM variants across base learners.}
        \label{fig:topk_comparison}
    \end{figure}

    We compare two APM variants: a deterministic top-$K$ variant and a probabilistic variant. Both follow a partial elitism idea that keeps exactly $K=5$ features per individual unchanged through the genetic operations: the deterministic variant always preserves the $5$ highest-importance features, whereas the probabilistic variant samples $5$ features without replacement, with each feature drawn with probability proportional to its importance. In both variants, features with zero importance are never preserved. As shown in \Cref{fig:topk_comparison}, the two variants are broadly equivalent, with the large majority of datasets showing no significant difference under either base learner. The minor differences that do arise vary by base learner, suggesting these differences stem from how feature importance is estimated under each base learner rather than a systematic advantage of either formulation. Overall, both variants perform similarly, confirming that feature-importance-guided protection is the key mechanism regardless of whether the protected features are selected deterministically or sampled probabilistically.

\end{document}